\pdfoutput=1

\documentclass[11pt]{article}

\usepackage[final]{acl}

\usepackage{times}
\usepackage{latexsym}
\usepackage[T1]{fontenc}
\usepackage[]{todonotes}
\newcommand{\SideNote}[2]{\todo[color=#1,size=\small]{#2}}
\newcommand{\emily}[1]{\SideNote{red!40}{#1 --emily}}

\usepackage{tabularx}
\newcolumntype{L}[1]{>{\raggedright\arraybackslash}p{#1}}
\usepackage{multirow}
\usepackage{comment}
\usepackage{booktabs}
\usepackage{enumitem}
\usepackage{pdfpages}

\usepackage[utf8]{inputenc}
\usepackage{microtype}
\usepackage{longtable}
\usepackage{supertabular}
\setlist[enumerate]{nosep, topsep=1pt}
\setlist[itemize]{nosep,topsep=1pt}
\makeatletter
\renewcommand{\paragraph}{%
  \@startsection{paragraph}%
    {4}%
    {\z@}%
    {0.5ex \@plus .2ex \@minus .2ex}%
    {-1em}%
    {\normalfont\normalsize\bfseries}%
}

\title{On Measures of Biases and Harms in NLP}

\author{Sunipa Dev$^{1}$\thanks{~~equal contribution} \qquad
Emily Sheng$^{2*}$ \qquad
Jieyu Zhao$^{1*}$ \qquad 
Aubrie Amstutz$^{*}$ \AND 
Jiao Sun$^{3}$ \qquad
Yu Hou$^{3}$  \qquad 
Mattie Sanseverino$^{1}$\qquad
Jiin Kim$^{1}$\qquad
Akihiro Nishi$^{1}$\AND
Nanyun Peng$^{1,3}$\qquad
Kai-Wei Chang$^{1}$ \\ \\
$^{1}$University of California, Los Angeles,
$^{2}$Microsoft Research,
$^{3}$University of Southern California \\
}

\begin{document}

\maketitle

\begin{abstract}
Recent studies show that Natural Language Processing (NLP) technologies propagate societal biases about demographic groups associated with attributes such as gender, race, and nationality. To create interventions and mitigate these biases and associated harms, it is vital to be able to detect and measure such biases. While existing works propose bias evaluation and mitigation methods for various tasks, there remains a need to cohesively understand the biases and the specific harms they measure, and how different measures compare with each other. 
To address this gap, this work presents a practical framework of harms and a series of questions that practitioners can answer to guide the development of bias measures.
As a validation of our framework and documentation questions, we also present several case studies of how existing bias measures in NLP---both intrinsic measures of bias in representations and extrinsic measures of bias of downstream applications---can be aligned with different harms and how our proposed documentation questions facilitates more holistic understanding of what bias measures are measuring.
\end{abstract}

\section{Introduction}
As language technologies and their applications become more widely deployed in our society, there are also increasing concerns of the disparate impacts and harms these technologies have on different demographic groups~\citep{bolukbasi2016man,webster2018mind}. 
To address some of these concerns, a large body of work has emerged to discuss~\cite{gonen2019lipstick,bender2021dangers,blodgett2021stereotyping}, detect~\cite{bolukbasi2016man,nangia2020crows-pairs}, measure~\cite{Caliskan183,zhao2019gender,webster2018mind,li2020unqovering}, and mitigate~\cite{dev2019attenuating,ravfogel2020null,sun-etal-2019-mitigating,dev-etal-2021-oscar} the social biases encoded by NLP models. 

Several of these works include bias measures comprising of metrics and datasets to define and investigate social biases within the constructs of a specific NLP task, such as text classification or machine translation.
Though these works propose different approaches for measuring biases, there is often similarly a lack of explicit alignment to harms, as well as a lack of comparative understanding of 
the advantages and disadvantages between the different bias measures for various language tasks. 
As an example, for the task of coreference resolution, there are several measures investigating gender bias~\citep{zhao2018gender,rudinger2018gender,lu2020gender,webster2018mind,cao2020toward}. 
However, each measure is unique in either the targeted demographic groups, metrics, dataset sentence structures, or the definition of bias, all of which ultimately affect what harms are measured. 
A better understanding of bias measures ultimately enables better adaptation and deployment for specific use cases. 

This paper is motivated by two main goals.
The first goal is to define a practical framework for harms that is both theoretically-motivated and empirically useful for describing bias measures. 
We organize a framework that is motivated by concepts from social psychology and linguistics, and narrow down specific definitions and heuristics 
to tag normative notions of harm with which bias measures align.
Moreover, we illustrate the utility of this measure-harm alignment exercise with case studies that demonstrate how a measure might unknowingly conflate different harms, or how separate measures with nearly identical task definitions can actually measure very different harms.
The second goal is to define a collection of documentation questions around bias measures that helps others capture measure limitations and align operationalizations of ``biases'' to harms.
Documenting various attributes (e.g., considerations for targeted demographic groups and tasks, dataset limitations, bias metric definitions and motivations) of a bias measure can help practitioners better articulate harms, appropriate use cases, and limitations.
To achieve these goals, we organize a practical framework of harms, a tagged collection of $43$ existing bias measures and the associated harms, a set of documentation questions, and a collection of case studies.



\begin{table*}
    \centering
    \scriptsize
    
    \begin{tabularx}{\linewidth}{L{4.5em} L{14.5em} L{17em} X}
    \toprule
            \bfseries Task & \bfseries {Demographic Dimension} & \bfseries Bias Measure & \bfseries Harms Evaluated \\
    \midrule
    
        \multirow{5}{*}{\parbox{5em}{Coreference\\Resolution}} & \multirow{1}{*}{\parbox{14.5em}{\textbf{Gender} through identity terms}}  & \citet{webster2018mind} & \textbf{QoS} \\ 
         & & \citet{cao2020toward} & \textbf{Erasure, QoS} \\ \cmidrule{2-4}
         
         & \multirow{1}{*}{\parbox{14.5em}{\textbf{Gender} through occupations}} & \citet{zhao2018gender} & \textbf{Erasure, Stereo.} \\ 
         & & \citet{rudinger2018gender} & \textbf{Erasure, Stereo.} \\
         & & \citet{lu2020gender} & \textbf{Erasure, Stereo.} \\\cmidrule{1-4}
        
         \multirow{5}{*}{\parbox{5em}{Natural\\Language\\Inference}} 
         & \parbox{14.5em}{\textbf{Gender} through occupations} & \citet{dev2019measuring}  & \textbf{Stereo.} \\ \cmidrule{2-4}
         & \multirow{1}{*}{\parbox{14.5em}{\textbf{Nationality} through identity terms}} & \citet{dev2019measuring} & \textbf{Disparagement, Stereo.} through polar adj. \\ \cmidrule{2-4}
         & \multirow{1}{*}{\parbox{14.5em}{\textbf{Religion} through identity terms}} & \citet{dev2019measuring} & \textbf{Disparagement, Stereo.} through polar adj. \\
        \cmidrule{1-4}
         
         \multirow{8}{*}{\parbox{5em}{Sentiment\\Analysis}} & \parbox{14.5em}{\textbf{Age} through identity terms} & \citet{diaz2018addressing} & \textbf{Disparagement, Erasure, QoS} through neg. sentiment \\ \cmidrule{2-4}
         & \parbox{14.5em}{\textbf{Gender} through identity terms} & \citet{kiritchenko2018examining} & \textbf{Dehumanization, Erasure, QoS, Stereo.} through emotion words \\ \cmidrule{2-4}
         & \parbox{14.5em}{\textbf{Rigid designators} through references to specific people} & \citet{prabhakaran2019perturbation} & \textbf{QoS} \\ \cmidrule{2-4}
         & \parbox{14.5em}{\textbf{Race} through identity terms} & \citet{kiritchenko2018examining} & \textbf{Dehumanization, Erasure, Stereo.} through emotion words \\
         \cmidrule{1-4}

         \multirow{7}{*}{\parbox{5em}{Question\\Answering}} & \parbox{14.5em}{\textbf{Race} through identity terms} & \citet{li2020unqovering} & \textbf{Erasure, Stereo.} through neg. assoc. \\ \cmidrule{2-4}
         & \parbox{13em}{\textbf{Ethnicity} through identity terms} & \citet{li2020unqovering} & \textbf{Erasure, Stereo.} through neg. assoc. \\ 
         & & \citet{li2020unqovering} + \citet{zhao2021ethical} & \textbf{Erasure, Stereo.} through neg. assoc. \\ \cmidrule{2-4}
         & \parbox{14.5em}{\textbf{Gender} through occupations} & \citet{li2020unqovering} & \textbf{Erasure, Stereo.} \\ 
         & & \citet{li2020unqovering} + \citet{zhao2021ethical} & \textbf{Erasure, Stereo.} \\ \cmidrule{2-4}
         & \multirow{2}{*}{\parbox{14.5em}{\textbf{Religion} through identity terms}} & \citet{li2020unqovering} &  \textbf{Erasure, Stereo.} through neg. assoc. \\
          & & \citet{li2020unqovering} + \citet{zhao2021ethical} & \textbf{Erasure, Stereo.} through neg. assoc. \\
         \cmidrule{1-4}

         \multirow{4}{*}{\parbox{5em}{Relation\\Extraction}} & \parbox{14.5em}{\textbf{Gender} through hypernym (occupation) relation} & \citet{gaut2019towards} & \textbf{Erasure, Stereo.} \\ \cmidrule{2-4}
         & \parbox{14.5em}{\textbf{Gender} through spouse relation} & \citet{gaut2019towards} & \textbf{Erasure, Stereo.} \\
         \cmidrule{1-4}
         
         \multirow{5}{*}{\parbox{5em}{Text\\Classification}} & \multirow{1}{*}{\textbf{Gender} through occupations} & \citet{De-artega2019bias} & \textbf{Erasure, Stereo.}\\ 
         & & \citet{zhao2020gender} & \textbf{Erasure, Stereo.}\\ \cmidrule{2-4}
         & \textbf{Gender} through identity terms & \citet{chalkidis2022fairlex} & \textbf{QoS} \\
         & \textbf{Age} through identity terms & \citet{chalkidis2022fairlex} & \textbf{QoS} \\
         & \textbf{Region} through identity terms & \citet{chalkidis2022fairlex} & \textbf{QoS} \\ \cmidrule{1-4}
         
         \multirow{30}{*}{\parbox{5em}{Toxicity\\Detection}} & \textbf{Age} through identity terms & \citet{dixon2018measuring} & \textbf{Disparagement, Erasure}\\
         &  &  \citet{sap2020social} & \textbf{Dehumanization, Disparagement, Erasure, Stereo.}\\ \cmidrule{2-4}
         & \textbf{Disability} through identity terms & \citet{dixon2018measuring} & \textbf{Disparagement, Erasure} \\
         &  & \citet{jigsaw} & \textbf{Disparagement, Erasure} \\
         &  &  \citet{sap2020social,hutchinson-etal-2020-social} & \textbf{Dehumanization, Disparagement, Erasure, Stereo.} \\ \cmidrule{2-4}
         & \textbf{Gender} through identity terms & \citet{dixon2018measuring} & \textbf{Disparagement, Erasure}\\
         & & \citet{park2018reducing}  & \textbf{Disparagement} \\
         &  & \citet{jigsaw} & \textbf{Disparagement, Erasure} \\
         &  & \citet{sap2020social} & \textbf{Dehumanization, Disparagement, Erasure, Stereo.} \\ \cmidrule{2-4}
         
         & \textbf{Rigid designators} through references to specific people & \citet{prabhakaran2019perturbation} & \textbf{QoS} \\ \cmidrule{2-4}
         
         & \textbf{Sexual Orient.} through identity terms & \citet{dixon2018measuring} & \textbf{Disparagement, Erasure} \\
         &  & \citet{jigsaw} & \textbf{Disparagement, Erasure} \\
         &  & \citet{sap2020social} & \textbf{Dehumanization, Disparagement, Erasure, Stereo.} \\ \cmidrule{2-4}
         & \textbf{Race} through identity terms &  \citet{dixon2018measuring} & \textbf{Disparagement, Erasure}\\
         &  & \citet{jigsaw} & \textbf{Disparagement, Erasure}\\
         &  &  \citet{sap2020social} & \textbf{Dehumanization, Disparagement, Erasure, Stereo.} \\ \cmidrule{2-4}
         & \textbf{Religion} through identity terms &  \citet{dixon2018measuring} & \textbf{Disparagement, Erasure}\\
         &  & \citet{jigsaw} & \textbf{Disparagement, Erasure}\\
         &  & \citet{sap2020social} & \textbf{Dehumanization, Disparagement, Erasure, Stereo.} \\ \cmidrule{2-4}
         & \textbf{Political Ideo.} through identity terms & \citet{sap2020social} & \textbf{Dehumanization, Disparagement, Erasure, Stereo.}
        \\ \cmidrule{2-4}
        & \textbf{Victim} through identity terms &  \citet{sap2020social} & \textbf{Dehumanization, Disparagement, Erasure, Stereo.} \\
         
    \bottomrule
    \end{tabularx}
     
    \vspace{-0.5em}
    \caption{Existing bias measures (part 1) organized by tasks, and demographic dimensions. A `+' indicates that one work built a bias metric (after `+') on top of a dataset from another work (before `+'). 
    \textit{Rigid designators}: references to specific people, \textit{polar adjectives}: good vs bad; \textit{negative activity}: violent or bad traits and activities. Sec.~\ref{sec:case-studies} delves into a few of these measures in the context of harms evaluated.
    } 
    
    \label{tbl: categories}
    \normalsize
\end{table*}

\section{Background}
\label{sec:definitions}

We clarify the definitions of several terms used throughout this paper.

\paragraph{Bias in NLP}
Bias in language models is commonly defined as ``skew that produces a type of harm'' \citep{crawford2017bias} towards different social groups, though it is a complex notion that is often not well-defined in existing literature~\citep{blodgett-etal-2020-language,delobelle-etal-2022-measuring,talat-etal-2022-reap}.
In the existing NLP literature, ``biases'' are often operationalized via a measurement model \citep{jacobs2021measurement} through bias measures.
While these bias measures are proxies for evaluating bias, they are often necessarily localized to measuring very specific skews and lack context of how a system would be used by real users.
Additionally, unstated assumptions and definitions often pervade these measures \citep{blodgett2021stereotyping}.
It remains an open question whether these bias measures actually measure meaningful and useful distinctions of ``biases''---this work provides initial explorations to answer this question for several measures.

\paragraph{Bias Measures}
Bias evaluations in NLP typically have been categorized broadly into \emph{intrinsic} or \emph{extrinsic} evaluations based on whether they measure biased associations within the word embedding spaces~\citep{Caliskan183} or biased decisions from models for specific tasks~\citep{mohammad2018obtaining,webster2019gendered}, respectively. 
We define a \emph{bias measure} 
as an evaluation standard that includes a \emph{metric(s)} applied to a \emph{dataset}.
Here, we use the term \emph{dataset} broadly, such that it could be applicable to datasets ranging in size and curation technique (e.g., manually crafted, generated).
To show inequalities between demographic groups, existing works typically define bias metrics (e.g., 
specialized notions of group fairness) that they then apply to a dataset specially designed to reveal social inequalities or stereotypes. 

These measures span several NLP tasks such as question answering~\citep{li2020unqovering}, relation extraction~\citep{gaut2019towards}, textual entailment~\cite{dev2019measuring}, toxicity prediction~\citep{dixon2018measuring,jigsaw,sap2020social}, coreference resolution~\citep{zhao2019gender,cao2020toward}, autocomplete generation~\citep{sheng2019woman}, dialogue generation~\citep{dinan2019build}, machine translation~\citep{stanovsky2019evaluating}, as well as intrinsic measurements of the embeddings themselves~\citep{Caliskan183,bolukbasi2016man,lauscher2020araweat,malik-etal-2022-socially}.

\paragraph{Demographic Dimension} 
We use the term \emph{demographic dimension} to refer to an identity axis (e.g., gender, race, age) for which specific instances (e.g., for gender: \textit{male}, \textit{female}, \textit{non-binary}, etc) are evaluated.
Instances of a demographic dimension are typically comparatively evaluated in measures through some proxy, e.g., occupations or identity terms.

\paragraph{Harms}
While existing works have examined possible harms of NLP models from various perspectives (e.g., general social impacts \citep{hovy2016social}, risks associated with large language models \citep{bender2021dangers}), in the context of algorithmic biases, we seek to align specifically with harms that can arise specifically from biases.
The relevant harms can be subdivided into representational or allocational harms, depending on whether there is a generalization of harmful representations of groups or if there is a tangible, disparate distribution of resources between groups, respectively~\citep{crawford2017bias}.\footnote{\citet{sheng2021societal} also separate out vulnerability harms, e.g., from model generations that render a group more susceptible to representational or allocational harms.}
In the context of aligning bias measures with targeted representational harms, one could align with the motivations for creating the measure (either explicit or unstated), the techniques used, or some mix of both. \citet{blodgett-etal-2020-language} present a categorization of the motivations and techniques of existing works that align with coarse-grained types of harms (\textit{allocational}, \textit{stereotypes}, \textit{other representational harms}), and \citet{blodgett2021sociolinguistically} further organize a taxonomy of fine-grained representational harm categories, including \textit{quality of service}, \textit{stereotyping}, \textit{denigration and stigmatization}, \textit{alienation},  and \textit{public participation}.
We build upon \citet{blodgett2021sociolinguistically}'s discussions, framing and extending our curated framework of harms through documentation questions and heuristics that can serve as a practical guide for those developing bias measures that capture specific harms.

Specifically, we use definitions of harms that are robust enough to capture aspects of a bias measure (dataset, metric(s), motivations) that align with different harms. Taking both individual and aggregate harms \citep{blodgett2021sociolinguistically} into consideration, this framework assumes vulnerability to harm is mediated by a dominant—non-dominant identity group dichotomy (inspired by but not entirely aligned with Social Dominance Theory~\cite{sidanius2001social}), which is helpful for operationalization purposes. 

In this paper, we focus on five types of harm: Stereotyping, Disparagement,\footnote{We choose to use ``Disparagement'' instead of ``Denigration'', to avoid invoking the conceptual metaphor of `blackening' one's reputation, which can have racial connotations in US culture.} 
Dehumanization, Erasure, and Quality of Service (QoS).
While there are other types of harms, and the five we target could be further broken down into subcategories, we start with these five as they are previously studied concepts and provide interesting insights to the non-exhaustive list of bias measures we examine in Table~\ref{tbl: categories} and Appendix Table~\ref{tbl: categories2}. 

\section{A Framework for Harms}
\label{sec: harm framework}
Conflating harms impedes accurate measurement; adequate and consistent delineation of harms enables ongoing appraisal of the effectiveness of mitigation strategies and the comparison of trade-offs.
Our practical framework of harms builds upon existing taxonomies of representational harms (e.g. ~\citet{blodgett2021sociolinguistically} and establishes specific heuristics (Appendix~\ref{sec:appendix-heuristics}) to disentangle the characteristics of five non-mutually exclusive categories. 
While these harms have previously been taxonomized, \textit{we ground the definitions of harms into documentation questions and heuristics to help practitioners align NLP bias measures with specific harms}.

Addressing a single phenomenon with different lenses can surface multiple harms; precisely which harm a method captures is sometimes solely dependent on the experimental framing, rather than some inherent taxonomic difference.
Using the harm heuristics we devise in Appendix~\ref{sec:appendix-heuristics}, we tag and distinguish between types of harms targeted by popular NLP bias measures presented in Tables~\ref{tbl: categories} and Appendix Table~\ref{tbl: categories2}. 
We note however, that other interpretations of targeted harms are certainly possible.  
This subjectivity makes it more crucial that those who build bias measures clearly state their motivations and include explanations of the relevant harms (Section~\ref{sec: questions}).\footnote{We also note that it is sometimes difficult to align with certain harms like Dehumanization without a closer examination of all samples in a measure dataset.}

\subsection{Harms}
\paragraph {Stereotyping}
Stereotypes are overgeneralized beliefs about the personal attributes of an individual as determined by their demographic group membership. Stereotypes as entities are codified associations which are necessarily well-known within a given context \citep{devine1989stereotypes} and can be expressed in infinite (and multi-modal) ways. Stereotypes draw on commonly held generalizations to make \textit{a priori} judgements about groups. In human cognition, they are perpetuated through a process of discounting counter-evidence as exceptions to the rule, e.g. confirmation bias~\citep{allport1954nature,link-2001}. 
These associations can in turn lead to unintended ``affective reactions'' by the model---precisely the measurable signals from which practitioners can infer bias.

\paragraph {Disparagement}
Disparagement
encapsulates any behavior by a model which reinforces the notion that certain groups are less valuable than others and less deserving of respect (or resources). Commonly associated measures of disparagement include toxicity ratings and hate speech detection scores.

\paragraph {Dehumanization}
Dehumanization actively casts disfavored groups as ``others'' and aims to erase signs of shared humanity (e.g. emotions, agency, intelligence), thus suppressing opportunities for empathy with said ``out group'' by characterizing them as sub-human~\cite{markowitz-2020,HASLAM201625}. Dehumanization can therefore be challenging to measure directly, as instances of dehumanizing language or sentiments are often closely intertwined with Disparagement and Stereotyping. 

\paragraph {Erasure}
Erasure refers to the lack of adequate representation of members of a particular social group~\cite{dev-etal-2021-harms,blodgett2022responsible}, whether intentional or not. While the data used to represent the intricacies of reality will always be necessarily incomplete, Erasure can arise from mismatches in reality and the data chosen to represent it. It can also serve to reinforce existing power structures via incautious mathematical averaging or aggregation of disparate groups. While relational group sizes from the real world can be reflected from the model in a quantitative sense, the challenge is designing systems which do not allow relative size to inappropriately affect prominence, i.e., attention needs to be paid to the potential effects these probabilities have on produced output.

\paragraph {Quality of Service}
Quality of Service harms result from instances where a model fails to perform equitably for different groups~\citep{blodgett2021sociolinguistically}. This harm can in turn potentially result in inequitable allocation of resources \citep{blodgett-etal-2020-language}, though this harm can also exist independently. The potential `quality' of service is operationalized and quantified via defined performance indicators, which can be systematically compared between commensurable groups.

\subsection {Relationships between Harms}
Disentangling which categories of harm a given bias measure measures requires careful articulation of the hypothesis and documentation of operationalization decisions; framing is crucial for producing substantively valid  results~\cite{jacobs2021measurement}. 
For example, an instance of bias in model training data may have arisen due to multiple types of harm or cause multiple types of harm. Our framework emphasizes how consequential these distinctions in operationalization can be.

\paragraph{Disparagement and Stereotyping}
Because stereotypes need to be codified and well-known within a given culture~\citep{devine1989stereotypes}, Disparagement is more generic and group-agnostic than Stereotyping. Consequently, datasets that test for Disparagement (explicitly or not) may sometimes be generated \textit{ad infinitum} by swapping demographic identifiers, e.g., ``[demographic identifier] are the worst kind of people''.
In comparison, the specificity required of statements expressing stereotypes presents limitations on rephrasing concepts (by design, languages have few ``absolute synonyms''~\cite{murphy_2010}).

\paragraph{Dehumanization, Disparagement, and Stereotyping
} 
Under our framework, Dehumanization contributes to Disparagement because it reinforces the idea that certain groups are inherently less valuable to society, i.e., Dehumanization always serves Disparagement, but not \textit{vice versa}. Dehumanizing language uses techniques such as \textit{moral disgust, denial of agency,} or \textit{likening members of a target group to non-human entities}~\cite{markowitz-2020} to reinforce normative identities---often as indication of a biological hierarchy of `species' within humankind. Dehumanization can be ``expressed tacitly''~\cite{markowitz-2020}, e.g., when groups are not considered worthy of being included (via Erasure). 
While descriptive, proscriptive, or prescriptive stereotypes~\cite{koenig2018comparing,hall2019mosaic} may have originated from some quantitative or qualitative fact about societal norms~\cite{sidanius2001social}, stereotypes which dehumanize are more likely inherently unfounded, e.g., stereotypes perpetuating racist pseudoscience like eugenics.

\paragraph{Stereotyping and Erasure} Cognitive heuristics like categorization and prediction based on probability are part of human nature~\cite{tversky1974judgment,mervis1981categorization}; 
however, harm can arise when these associations obfuscate or erase actual variance (e.g., via confirmation bias) or when society assigns a cost (e.g., social, allocational) when these oversimplified ``norms'' are not adhered to by their respective group members (e.g., proscriptive or prescriptive~\cite{koenig2018comparing}). 
Erasure and Stereotyping can have a cyclical relation; lack of representation of variance and sub-populations can both result in stereotypes and be a direct result of Stereotyping. Erasure and Stereotyping are conceptualized as being one level of abstraction away from the consequence being caused: while exposure to a disparaging or dehumanizing remark can be directly harmful in the moment, the impact of Stereotyping associations and Erasure are more apparent at a distributional level. Additionally, Erasure and Stereotyping are strongly mediated by the \textit{vulnerability} of the group and the \textit{severity} of the implications of the association.

\paragraph{Quality of Service and Erasure}
Facts about historical inequities, social hierarchies, and stereotypes should guide Erasure measures. 
Under our framework, measures that target Erasure harms should have strong, directional hypotheses 
in order to surface representation issues for specific groups.
These issues could in turn be quantified more precisely via comparative evaluation methods, such as those common in measures that target Quality of Service harms. Erasure measurement for under-represented groups requires us to set aside quantitative majorities and ensure qualitative ``coverage'' instead, e.g., while there may be fewer female than male surgeons in the United States, the former do still exist. 
The desired effect of removing Erasure harms is for representation of actual diversity to persist, independent of statistical presence. 

\section{Documenting Bias Measures}
\label{sec: questions}
While bias measures aimed at various tasks are widely developed across the NLP community, the measures are often underused or re-developed by researchers for the same task. This stems largely from a lack of usability since little to no documentation of motivation and various choices is available for these measures. Documentation for datasets and models have proliferated over the last few years but the rapidly growing collection of bias measures lacks such organized efforts. 


Existing works have stressed the importance of documenting models \citep{mitchell2019model}, datasets \citep{gebru2018datasheets,bender2018data}, measurement modeling validity and reliability~\citep{jacobs2021measurement}, and, more recently, ethical considerations \citep{mohammad-2022-ethics}.
This paper adds a complimentary resource focusing on documentation considerations for bias measures into the existing collection.
In this work, we build upon the existing guidelines from \citet{gebru2018datasheets}, which are more generally for datasets of any modality or purpose, and 
narrow the focus to bias measures for NLP tasks.
We add questions related to the \emph{Composition} and \emph{Collection Process} sections as proposed by \citet{gebru2018datasheets}. Additionally, we propose new sections on \emph{Motivation} specifically for bias measures and \emph{Bias Metrics}. 
The specificity of the questions addresses the intended usage of different bias measures more explicitly.

\begin{enumerate}[leftmargin=*]
    
\item \textit{Motivation}

\citet{blodgett-etal-2020-language} detail the importance of concretely defining the biases being measured and listing out how a metric aligns with normative definitions of harm. Additionally, discerning biases from model errors is equally important and particularly ambiguous when a definition for the ``bias'' measured is absent.

\begin{itemize}[leftmargin=*]

    \item \textbf{What is the stated definition of bias?}
    
    \item \textbf{How does this definition align with normative definitions of harm?}
    For a measure to be a valid quantification of bias, the notion of ``bias'' has to be well-defined and related to what is measured.
    More explicitly bridging the gap between bias metrics and 
    harms can tangibly disambiguate between innocuous model errors and potential harms downstream.
    \item \textbf{If the bias measure measures more than one harm, are the harms conflated in one measurement or separable?} A single instance of language may represent/cause multiple forms of harm (e.g., some Stereotyping harms may also be Dehumanization harms). Does the measure provide a method for measuring multiple harms separately as well as in aggregate (e.g., are subsets of the underlying data tagged along multiple axes)? 
    \item \textbf{What language and culture is the bias and measure most relevant to?}
    \item \textbf{
    What other contexts can the measure be extended to?} This question is intended to obtain a list of the specific demographic groups and locales a bias measure has been shown to be useful for.
    \item \textbf{If a demographic attribute is split into groups for measurement of bias, how many groups have been considered?} What is the justification for the grouping? Have prominent/consequential intersectional identities been considered? This question is to understand the scope of the measure and assess its coverage.
    \item \textbf{What is the source of bias that is measured?} Social biases creep into NLP models in different ways - the data used to derive representations, the model (and parameters) used, etc. The bias measured can be from one or all sources and needs to be acknowledged and when possible, disambiguated.
    \item \textbf{What tasks or applications is this bias measure useful for?} Is this measure effective to check on any language representations for social biases irrespective of application? Or is there a specific task where this is most applicable?
\end{itemize}

\item \textit{Composition and Collection Process}

Language data for bias measures is sourced primarily in two ways:
by extracting from existing textual data or by generating from specific templates. While the first has the advantage of being more similar to ``real samples'' that models see, the latter has the advantage of testing for specific artifacts by construct.

\begin{itemize}[leftmargin=*]
    \item \textbf{Is the bias measure data scraped, generated, or produced some other way?} Scraping or generating text using templates are two common ways of building bias measure datasets in NLP, and different dataset curation techniques have their own advantages and disadvantages.
    \item \textbf{What are the limitations associated with method of data curation? How generalizable is this dataset? 
    } 
    Examples of limitations include scraped English text containing predominantly Western narratives and data annotated by annotators with specific biases.
    \item \textbf{If the dataset is scraped, what are the primary sources/domains?} Some text sources are known to harbor more toxic or harmful content than others.
    \item \textbf{What is the structure of the sentence, sentence segment, template, or trigger phrase used for data collection?} Does the particular structure come with certain simplifications, assumptions, or guarantees?

    \item \textbf{Is the dataset at risk of causing harm through the particular selection of proxy attributes representing demographic groups?} For example, does this dataset use popular names as a proxy for gender? Is there a risk for misidentifying individuals if the associated genders are not self-reported? Does the expected gender - name pairing align with the time period of the sourced data? 
\end{itemize}

\item \textit{Bias Metrics}

This section presents documentation questions for metrics that are used with datasets to measure bias.
Specific definitions and comparisons can broaden understanding about the measured biases.
\begin{itemize}[leftmargin=*]
    \item \textbf{How is the bias metric defined? Is there a null hypothesis or normalization recommended for it to be meaningful?}
    \item \textbf{Is it an absolute or relative evaluation?} 
    \citet{sheng2021societal} describe absolute score evaluations as those that ``use an accumulated score to summarize inequalities between
demographics, whereas relative evaluations explicitly report inequalities between all demographics.'' Absolute scores offer more simplicity, and relative scores offer more flexibility in alignment with normative harms.
    Through this question, we hope to understand the motivation behind the evaluation format.
    \item \textbf{Are there alternate or existing metrics this metric can or should be used with?} This question covers the cases where a bias metric may not be enough to measure all desired metric attributes, either in terms of bias or general task evaluations. 
    \item \textbf{Are there other existing datasets or metrics to evaluate bias for the same task?} How does an evaluation using one metric correlate with another using a different metric?
    Note that high correlation between measures do not necessarily imply meaningful or useful measures.
    Additionally, does the sentence structure, sourcing method, or other feature differ between the datasets? 
    \item \textbf{Can the metric imply an absolute absence of bias in a specific task or model?} Are there other measurements needed for a complete assessment of bias? Is a complete assessment possible?
\end{itemize}
\end{enumerate}

\section{Case Studies}
\label{sec:case-studies}
We present a series of case studies as examples of how our proposed framework of harms and documentation questions reveal unique insights into different bias measures. In Table~\ref{tbl: categories} and Appendix Table~\ref{tbl: categories2}, we tag bias measures with the relevant, targeted harm(s). In this section, we discuss concrete examples to elucidate how subtle differences in framing of measures impact the harm(s) measured.

\subsection{Disparagement and Stereotyping}
To better understand the subtleties between Disparagement and Stereotyping, we examine two existing bias measures.

\citet{davani2020fair} present a fair hate speech measure that implicitly separates Stereotyping and Disparagement harms; however, these alignments are not explicitly connected, and our framework helps distinguish between the two harms.
This work of \citet{davani2020fair} is motivated by the observation that not all demographic groups are interchangeable when it comes to specific stereotypes.
For example, they note that substituting ``Muslim'' with ``Jew'' in a hateful sentence about terrorism does not create equivalently valid stereotypes within the US cultural context.
Thus, they create ``symmetric counterfactual'' statements that convey a similar meaning when different group tokens are substituted.
Interestingly, this distinction between symmetric and asymmetric counterfactuals helps delineate between Disparagement and Stereotyping sentences, as symmetric counterfactuals are, by nature, generic enough to disparage multiple groups. Unless two independent stereotypes have coincidentally converged (e.g., two groups are associated with terrorism for different historical reasons within a given context), a carrier phrase that is able to substitute group identifiers is unlikely to be able to produce valid stereotypical sentiments.
Thus, this process of creating and making the distinction between symmetric and asymmetric counterfactual tests generates a fair hate speech dataset that includes some amount of coverage for both Disparagement and Stereotyping harms.



\citet{dev2019measuring} is another example where Disparagement and Stereotyping harms are not explicitly separated.
This work measures biases in the task of natural language inference by
comparing demographic associations with polar adjectives.
We find that this particular setup conflates Disparagement and Stereotyping harms. 
As an example from the dataset, for the template ``[demographic identifier] are [adjective]'', 
the statement ``Canadians are nice'' is a stereotype, whereas another statement such as ``Uzbekistanis are bad'' is more of a general disparaging remark than a stereotype.

These examples show the difficulty in carefully designing datasets that test for Stereotyping versus Disparagement harms.

\subsection{Quality of Service, Stereotyping, and Erasure}
Next, we present an empirical case study examining 
how measures designed for the same task can differ in the harms measured.
\citet{webster2018mind} and \citet{cao2020toward} both discuss biases in the task of coreference resolution where the goal is to identify phrases or terms referring to the same entity in a sentence.
\citet{webster2018mind} measure biases in the model's ability to correctly resolve gendered 
pronoun-name relationships for the binary genders and
is aligned with the Quality of Service harm, since the measure 
probes the contrastive relationship between model performance for females versus males.  
\citet{cao2020toward} expand the \texttt{GAP} dataset introduced by \citet{webster2018mind} to create the \texttt{MAP} dataset, where the authors swap out gendered words for a set of gender neutral variations of the sentences in \texttt{GAP}.
While both \texttt{GAP} and \texttt{MAP} are part of bias measures that are aligned with Quality of Service harms, \texttt{MAP} also surfaces Erasure harms by testing for whether a coreference system fails to process text for non-binary pronouns. 

Additionally, two other popularly used bias measures for coreference resolution, as described by \citet{rudinger2018gender} and \citet{zhao2019gender}, compare the association of specific occupations with gendered pronouns. While some dataset instances directly measure Stereotyping harms, such as a preferential association of `doctor' with typical male pronouns, other instances do not directly measure explicit stereotypes in the society but rather an implicit Erasure or lack of representation of some genders in overall text. 
While both of these harms are overall conflated by both measures, unlike \texttt{GAP} and \texttt{MAP}, neither measures Quality of Service harms. 

\subsection{Dehumanization and Stereotyping}
\citet{kiritchenko2018examining}'s bias measure for sentiment analysis formulates a dataset of simple sentences including names, gendered pronouns, and other indicators of demographic group identity, and compares the sentiment associated with different groups. While some sentences evaluate stereotypes such as the ``Angry Black Woman'', others are not indicative of any stereotype but rather analyze the societal license for a member of a certain group to display a range of emotions--i.e., Dehumanization.
The two harms measured are not distinguishable by the metric used, but instead by careful examination of the individual sentence templates, word lists, and names used. 


 
\subsection{Insights from Documenting Bias Measures}
By using our harm framework to label the bias measures in Table~\ref{tbl: categories} and Appendix Table~\ref{tbl: categories2} as well as documenting bias measure motivations and compositions, we developed several insights.

The first is that \textit{documentation facilitates deeper analysis and should be revisited periodically}.
We use the proposed questions to analyze the work described by \citet{sheng2019woman}.
In particular, we note that there is no explicit definition of biases in the work, although the operationalization of their regard metric as a measure of social perception aligns with the measurement of representational harms (e.g., Stereotyping and Disparagement).
In answering the documentation questions (Appendix~\ref{appendix:case-study-regard}), we find that this documentation exercise is especially useful if the documented measure has been released for a while.
In the case of the regard metric of \citet{sheng2019woman}, there were not many points of comparison at the time of its release, but more relevant comparisons have recently been released.
Thus, we recommend treating documentation as a continuous process and revisiting the questions regularly.

Also, \textit{documentation reveals specific limitations across bias measures for a specific task}.
The specificity of the documentation questions helps uncover what is currently measured and encourages the development and use of complementary measures.
In documenting WinoBias~\citep{zhao2018gender} in Appendix~\ref{appendix:case-study-winobias},
we examine various bias measures for coreference resolution more closely.
Existing bias measures for coreference resolution that target gender biases through occupations have all focused on associated stereotypes and the relative representation between binary genders, and thus target Stereotyping and Erasure harms, as shown in Table~\ref{tbl: categories}.
On the other hand, the coreference resolution bias measures that target gender through identity terms explore the effect of model performance for gender-neutral pronouns, and thus target Quality of Service (and some Erasure) harms.

A third insight is that \textit{inherent constraints of a task seem to affect the method by which bias measures (implicitly or explicitly) target harms}. For more constrained language understanding tasks in which the model produces a limited set of outputs (e.g., classification), the dataset designed for the measure largely affects the targeted harm. 
For example, for measuring biases in coreference resolution, the standard metrics are $F_1$ or accuracy scores---it is really by examining the datasets (and motivations) that we discern whether we are targeting Stereotyping (e.g., through occupational associations) or Quality of Service harms. For open-domain language generation tasks, targeted harms are largely affected by the selected bias metrics rather than the datasets. 
Because generation task are so open-ended, it is often difficult to design evaluation datasets that achieve a lot of control over the resulting model output, and thus existing works rely more on various bias metrics to capture different harms. 
For example, \citet{dhamala2021bold} evaluate biases using sentiment, regard, toxicity, and psycholinguistic norms to target different operationalizations of harms.

\section{Conclusion}
Bias measures in NLP are critical for estimating and mitigating potential harms towards different demographic groups. 
However, a lack of structured understanding of what harms exist, how they are operationalized through bias measures, and how they can be measured can diminish the usefulness of bias measures. 
In this work, we organize a framework to define and distinguish between different types of harms---presented through heuristics and documentation questions---to guide more intentional development of bias measures.
Our proposed documentation template also facilitates 
combining, comparing, and utilizing different bias measures, and 
continuously re-visiting them to update limitations and comparative understanding with other measures.

\section{Limitations and Ethical Considerations} 
We acknowledge that our framework of harms has been created from a US-centric perspective and has been influenced by the Social Dominance Theory \citep{sidanius2001social}, which can be limiting from a global perspective and does not include cultural harms. While some definitions and operationalizations of harms in our framework (e.g., Stereotyping, Disparagement) may be applicable to other cultural perspectives, we note that there may be some that require cultural context-specific updates and also that there are other harms that we did not cover. There are also other bias measures in this rapidly growing space that we may not have covered and tagged with harms measured.
Additionally, we do not focus on specific downstream applications where each measure might be used and encourage further analysis on these applications.

We further emphasize that while documentation enables more transparency into bias measures, documentation \textit{does not ensure the validity} of the measures.
In fact, there is a risk that the act of documentation could give a measure a false sense of validity.
Too many documentation questions may also become an obstacle for practitioners interested in working on a topic, though we believe it is better for community progress to start thinking about these questions before designing bias measures.

\section*{Acknowledgments}
We would like to thank various people for their valuable discussion and feedback, including
Alexandra Olteanu, Arjun Subramonian, Chad Atalla, Dan Vann, Emily Corvi, Hanna Wallach, Hannah Washington, Jason Teoh, Kevin Robinson, Stefanie Reed, Su Lin Blodgett, Vinodkumar Prabhakaran, as well as our anonymous reviewers. This work was supported by NSF grant \#1927554, NSF grants
\#2030859 and \#2127309 to the Computing Research Association for the CIFellows Project, and the Sloan Award.

\newpage
\bibliography{custom,nlp,ijcai22,ref}

\clearpage
\newpage

\pagenumbering{arabic}

\twocolumn[
\begin{center}
{\Large \textbf{\\ Appendix: On Measures of Biases and Harms in NLP\\ \vspace{1in}}}
\end{center}
]

\appendix

\begin{table*}[!t]
    \centering
    \scriptsize
    
    \begin{tabularx}{\linewidth}{L{4.5em} L{15.5em} L{24em} X}
    \toprule
        \bfseries Task & \bfseries {Demographic Dimension} & \bfseries Bias Measure & \bfseries Harms Evaluated \\
    \midrule
    
    \multirow{20}{*}{\parbox{5em}{Hate \\Speech\\Detection}} & \textbf{Gender} through identity terms & \citet{davani2020fair} & \textbf{Disparagement, QoS, Stereo.} \\ \cmidrule{2-4}
        & \textbf{Gender} through stereotypes & \citet{founta2018large} + \citet{goldfarb2020intrinsic} & \textbf{Disparagement} \\
        &  &  \citet{basile2019semeval-2019} + \citet{goldfarb2020intrinsic} & \textbf{Dehumanization, Disparagement} \\ \cmidrule{2-4}
        & \textbf{Migrants} through identity terms & \citet{davani2020fair} & \textbf{Disparagement, QoS, Stereo.} \\ \cmidrule{2-4}
        & \textbf{Migrants} through identity terms & \citet{basile2019semeval-2019} + \citet{goldfarb2020intrinsic} & \textbf{Dehumanization, Disparagement} through pleasantness terms \\ \cmidrule{2-4}
        & \textbf{Political Ideo.} through identity terms & \citet{davani2020fair} & \textbf{Disparagement, QoS, Stereo.} \\ \cmidrule{2-4}
         & \textbf{Race} through dialect & \parbox{24.5em}{[\citet{blodgett2016demographic}, \citet{ davidson2017automated}, \citet{founta2018large}, \citet{preotiuc-pietro-ungar-2018-user}] + \citet{sap2019risk}} & \textbf{Disparagement, Erasure, QoS} \\
         &  & \parbox{24.5em}{[\citet{blodgett2016demographic}, \citet{ davidson2017automated}, \citet{founta2018large}] +  \citet{xia2020demoting}} & \textbf{Disparagement, Erasure} \\
         & & \parbox{24.5em}{[\citet{waseem2016hateful}, \citet{waseem-2016-racist}, \citet{ davidson2017automated}, \citet{founta2018large}, \citet{golbeck2017harassment}, \citet{blodgett2016demographic}] + \citet{davidson2019racial}} & \textbf{Disparagement, Erasure, QoS} \\\cmidrule{2-4}
         
        & \textbf{Race} through identity terms & \citet{davani2020fair} & \textbf{Disparagement, QoS, Stereo.} \\
        & & \citet{kennedy2020contextualizing} & \textbf{Dehumanization, Disparagement}\\ \cmidrule{2-4}
        & \textbf{Religion} through identity terms & \citet{davani2020fair} & \textbf{Disparagement, QoS, Stereo.} \\ \cmidrule{2-4}
        & \textbf{Sexual Orient.} through identity terms & \citet{davani2020fair} &  \textbf{Disparagement, QoS, Stereo.}\\
         \cmidrule{1-4}
         
        
    \multirow{22}{*}{\parbox{5em}{MLM \\Predictions}} & \textbf{Age} through identity terms & \citet{nangia2020crows-pairs} & \textbf{Stereo.} \\
    & & \citet{neveol2022french} & \textbf{Stereo.} \\ \cmidrule{2-4}
    
         & \textbf{Appearance} through identity terms & \citet{nangia2020crows-pairs} & \textbf{Stereo.} \\
         & & \citet{neveol2022french} & \textbf{Stereo.} \\ \cmidrule{2-4}
         
         & \textbf{Disability} through identity terms & \citet{nangia2020crows-pairs} & \textbf{Stereo.} \\
         & & \citet{neveol2022french} & \textbf{Stereo.} \\ \cmidrule{2-4}
         
         & \textbf{Gender} through identity terms & \citet{nangia2020crows-pairs} & \textbf{Stereo.} \\
          &  & \citet{nadeem2020stereoset} & \textbf{Stereo.} \\
          & & \citet{neveol2022french} & \textbf{Stereo.} \\ \cmidrule{2-4}
          
         & \textbf{Nationality} through identity terms & \citet{nangia2020crows-pairs} & \textbf{Stereo.} \\
         & & \citet{neveol2022french} & \textbf{Stereo.} \\ \cmidrule{2-4}
         
         & \textbf{Race} through identity terms & \citet{nangia2020crows-pairs} & \textbf{Stereo.} \\
         &  & \citet{nadeem2020stereoset} & \textbf{Stereo.} \\
         & & \citet{neveol2022french} & \textbf{Stereo.} \\ \cmidrule{2-4}
         
          & \textbf{Religion} through identity terms & \citet{nangia2020crows-pairs} & \textbf{Stereo.} \\
         &  & \citet{nadeem2020stereoset} & \textbf{Stereo.} \\
         & & \citet{neveol2022french} & \textbf{Stereo.} \\ \cmidrule{2-4}
         
          & \textbf{Sexual Orient.} through identity terms & \citet{nangia2020crows-pairs} & \textbf{Stereo.} \\
          & & \citet{neveol2022french} & \textbf{Stereo.} \\ \cmidrule{2-4}
          
         & \textbf{Socioeconomic} through identity terms & \citet{nangia2020crows-pairs}& \textbf{Stereo.} \\
          & & \citet{nadeem2020stereoset} & \textbf{Stereo.} \\
          & & \citet{neveol2022french} & \textbf{Stereo.} \\
         \cmidrule{1-4}
         
    \multirow{13}{*}{\parbox{5em}{Autocomplete\\Generation}} & \textbf{Gender} through identity terms & \citet{sheng2019woman} & \textbf{Disparagement, Stereo.} \\
          &  & \citet{huang2020reducing} & \textbf{Erasure, Stereo.} \\ 
          &  & \citet{dhamala2021bold} & \textbf{Disparagement, Stereo.} \\ \cmidrule{2-4}
          & \textbf{Gender} through occupations & \citet{alnegheimish2022using} & \textbf{Erasure, Stereo.} \\ 
          \cmidrule{2-4}
         & \textbf{Race} through identity terms & \citet{sheng2019woman} & \textbf{Disparagement, Stereo.} \\
          &  & \citet{dhamala2021bold} & \textbf{Disparagement, Stereo.} \\ \cmidrule{2-4}
         & \textbf{Race} through dialect & \citet{groenwold2020investigating} & \textbf{Erasure, Stereo.} \\ \cmidrule{2-4}
         & \textbf{Sexuality} through identity terms &  \citet{sheng2019woman} & \textbf{Disparagement, Stereo.} \\ \cmidrule{2-4}
         & \textbf{Country} through identity terms & \citet{huang2020reducing} & \textbf{Erasure, Stereo.} \\ \cmidrule{2-4}
         & \textbf{Occupation} through identity terms &  \citet{huang2020reducing} & \textbf{Erasure, Stereo.} \\
         & &  \citet{dhamala2021bold} & \textbf{Disparagement, Stereo.} \\ \cmidrule{2-4}
         & \textbf{Religion} through identity terms &  \citet{dhamala2021bold} & \textbf{Disparagement, Stereo.} \\ \cmidrule{2-4}
         & \textbf{Political Ideo.} through identity terms &  \citet{dhamala2021bold} & \textbf{Disparagement, Stereo.} \\
         \cmidrule{1-4}
         
         \multirow{3}{*}{\parbox{5em}{Dialogue\\Generation}} & \textbf{Gender} through identity terms &  \citet{liu2020does, liu2020mitigating} & \textbf{Disparagement, Stereo.}\\
         & & \citet{dinan2020queens} & \textbf{Dehumanization, Erasure, Stereo.} \\ \cmidrule{2-4}
         & \textbf{Race} through identity terms &  \citet{liu2020does} & \textbf{Disparagement, Stereo.}\\
         \cmidrule{1-4}
         
         \multirow{4}{*}{\parbox{5em}{Translation}} & \textbf{Gender} through occupations & \citet{stanovsky2019evaluating} & \textbf{Erasure, QoS, Stereo.}\\ \cmidrule{2-4}
         & \textbf{Gender} through identity terms & \citet{wang2022measuring} & \textbf{Erasure, QoS, Stereo.} \\
         & \textbf{Nationality} through identity terms & \citet{wang2022measuring} & \textbf{Erasure, QoS, Stereo.} \\
         & \textbf{Race} through identity terms & \citet{wang2022measuring} & \textbf{Erasure, QoS, Stereo.} \\
         \cmidrule{1-4}
         
        \multirow{2}{*}{\parbox{5em}{Text\\Re-writing}} & \textbf{Gender} through inflections & \citet{habash2019automatic} & \textbf{Erasure, Stereo.} \\
        & & \citet{zmigrod2019counterfactual} & \textbf{Erasure, Stereo.}\\
         
    \bottomrule
    \end{tabularx}
     
    \vspace{-0.5em}
    \caption{Existing bias measures (pt. 2) by tasks and demographics. `+' means that one work built a bias metric (after `+') on top of a dataset from another (before `+'). Brackets group datasets that were all used by a metric.
    } 
    
    \label{tbl: categories2}
    \normalsize
\end{table*}

\section{Harm Framework Heuristics}
\label{sec:appendix-heuristics}
To help practitioners determine the specific harm(s)
a bias measure evaluates, we propose the following set of heuristics.

\paragraph{Stereotyping:}
\label{appendix:stereotyping}
Does the method:
\begin{itemize}
\item deal with language which communicates an existing, well-known a priori judgement or generalization which oversimplifies the reality of diversity within the group?
\item measure predictions or probabilities of associations between specific groups and certain characteristics, concepts, language, or sentiments? 
\item focus on finding specific, pre-defined outcomes based on hypotheses about stereotypical associations, i.e., is the hypothesis directional?
\item test associations which either the "average" in-group member or person in the relevant society would be able to quickly predict, i.e., would they be able to predict or identify what the 'problem' is and connect its roots to their cultural/historical knowledge?
\end{itemize}
Note: these associations can be positive or negative, but should not hold as naturalistic when a commensurable group is swapped in.

\paragraph{Erasure:} 
Does the method:
\begin{itemize}
\item search for lack of representation of specific groups based on cultural trends and patterns of historical inequality?
\item engage with mismatches between representation and reality (due to imprecise categorizations, rounding errors, etc.)?
\item interrogate representation issues caused by prevailing stereotypes, dehumanization, or cultural narratives? 
\item primarily concern itself with whether or how specific, pre-defined groups are represented or treated equitably, rather than to what extent groups are treated inequitably in relation to one another?
\item primarily provide results about the model performance for a specific group in relation to a ‘control’ group (whether or not explicitly stated as such)? 
\end{itemize}

\paragraph{Disparagement:}
Does the method:
\begin{itemize}
\item deal with generally belittling, devaluing, or de-legitimizing language about a group?
\item engage with sentiments related to societal regard (respect), expressing normative judgments, or using scalar adjectives pertaining to quality or worth (best/worst, good/bad), but which are not tied to an established stereotype?
\item use language which holds as pragmatically and semantically valid/naturalistic when the group identifier is perturbed with a commensurable group?
\item deal with `toxicity' or `unhealthy' discourse in general?
\end{itemize}

\paragraph{Dehumanization:}
Does the method specifically mention language commonly used to dehumanize, such as:
\begin{itemize}
\item associations with non-human life (vermin, insects)?
\item implications that a certain group is sub-human or not ‘true’ members of a superset (certain ‘immigrants’ aren’t ‘American’)?
\item notions related to eugenics?
\item justifications of inequitable treatment of groups or denial of human rights based on group membership
(note: these can be codified into stereotypes, but are distinguished by their unique purpose to ‘other’ the group, reinforcing notions of normative identities and casting divergence as indication of a hierarchy of ‘species’ within humankind)?
\end{itemize}

\paragraph{Quality of Service: }
Does the method:
\begin{itemize}
\item seek to measure the comparative performance of a model for several commensurable demographic groups? 
\item have an obvious or direct application to mitigation efforts or industry usage?
\item primarily concern itself with to what extent groups are treated inequitably (quantification), rather than whether they are treated differently?
\end{itemize}

\section{A Survey of Bias Measures for Understanding Harms}
\label{sec:survey}
As NLP models grow in size, complexity, ability to mimic underlying languages, and the extent to which they are deployed in real world applications, it becomes more important to understand their potential for biases and harms.
A growing number of measures serve to evaluate biases 
in tasks such as sentiment analysis or relation extraction, targeting specific social biases related to gender, race, religion, etc. 
While measures to evaluate biases have been formulated across various tasks, there remains a lack of cohesive understanding of \emph{what these bias measures evaluate} 
and \emph{how different measures relate}. 
In this section, we survey and describe a non-exhaustive list of measures for quantifying biases in different NLP tasks for primarily English. 
Tables~\ref{tbl: categories} and \ref{tbl: categories2} summarize this survey along with alignments of harms for different bias measures.

\subsection{Natural Language Understanding}
We discuss existing works that use different measures to assess the presence of social biases in a variety of NLU tasks.
\paragraph{Coreference Resolution}
Coreference resolution is the task of finding all expressions that refer to the same entity in text; a more specific objective is to associate pronoun mentions to different entities. There are two distinct definitions of bias that are evaluated with respect to this task, both centered around gender.
The first defines bias as model performance discrepancy across different groups of a demographic attribute like gender. 
The Gendered Ambiguous Pronouns ($GAP$) dataset~\citep{webster2018mind} consists of samples from Wikipedia biographies with ambiguous pronoun-name resolution pairs.
\citet{webster2018mind} defines and measures biases through a disparity in correctly resolving pronoun-name relationships for the male and female genders. 
The Maybe Ambiguous Pronoun ($MAP$) dataset~\citep{cao2020toward} expands $GAP$ to go beyond binary genders with a broader dataset.
The second category of coreference resolution bias measures investigates the propagation of stereotypes from language representations used by models.
Both WinoBias~\citep{zhao2018gender} and Winogender~\citep{rudinger2018gender} generate Winograd schema style datasets to investigate occupational gender stereotypes.
Additionally, \citet{lu2020gender} create sentence templates to evaluate biases using the ratio of accurate pronoun resolution for stereotypical vs non-stereotypical occupational associations.
 
Existing works that use the second definition of bias currently focus on singular stereotypes (e.g., with regards to occupation), while gender biases can encompass a broad range of other stereotypical and undesired associations. 
While both definitions of bias can potentially cover additional demographics and undesired associations, it is important to question which is more applicable to investigate harms faced by a group. 
For example, non-binary individuals face erasure in language representations~\cite{dev-etal-2021-harms}, and these experienced harms might be more appropriately captured by the first definition, whereas stereotyping might be by the second.

\paragraph{Natural Language Inference (NLI)}
NLI determines the directional relationship between two sentences, as to whether the second sentence (hypothesis) is entailed, contradicted, or neutral to the first sentence (premise). \citet{dev2019measuring} demonstrate how the task captures and mirrors stereotypical associations (with binary gender, religion, etc) learned by text representations. 
Their bias measure consists of a dataset with sentence pairs: one sentence with an explicit demographic attribute (e.g., gender), and the other with implicit, stereotypical associations (e.g., occupations).  
Bias is measured as the accuracy of models in identifying that all sentences have no directional relation, i.e., classified as having the `neutral' label. 
Since an overall bias score is calculated over a set of templates, a variety of templates can be independently assessed together to evaluate bias of NLI model outcomes across multiple demographic groups, thus not restricting measurements to a single stereotype.

\paragraph{Sentiment Analysis}
Estimating the sentiment or language polarity of text is useful for understanding consumer perception from reviews, tweets, etc. However, this task has been demonstrated to be stereotypically influenced by demographic characteristics such as race and gender~\citep{kiritchenko2018examining}, age~\citep{diaz2018addressing} and names of individuals~\citep{prabhakaran2019perturbation}.
Existing works keep sentence templates constant between samples and change the assumed demographic attribute of the person (e.g., through names) in a sentence. This ideally should not change the sentiment classification of the sample---any changes in sentiment indicate the existence of stereotypical associations. Since evaluation hinges on this contrast in classification across groups, bias against a group is also measured in comparison to another.

\paragraph{Question Answering (QA)}
QA models perform reading comprehension tasks and also propagate stereotypical associations from underlying language representations, as demonstrated through UnQuover~\citep{li2020unqovering}. 
\citet{li2020unqovering} use sentence templates containing limited direct demographic information (e.g., names) and under-specified questions containing no related demographic information to measure biases exhibited by QA systems.  
The setup is such that all sub-categories of a demographic attribute (e.g., religion: Christian, Buddhist, etc) should be equally predicted as the answer. A statistically significant, higher value for one sub-category is interpreted as bias. 
Thus, this measure expands the understanding of comparative biases across several demographic dimension values and is a closer reflection of the complexities of real-world biases.

\paragraph{Neural Relation Extraction}
Relation extraction is the task of extracting relations between entities in a sentence and is instrumental in converting raw, unstructured text to structured data. \citet{gaut2019towards} note how gender biases in this task could lead to allocational harms by affecting predictions on downstream tasks. They create a dataset, WikiGenderBias, 
containing sentences regarding either a male or female entity and one of four relationships: spouse, occupation, birth date, or birth place. 
Similar to $GAP$, the evaluation framework measures gender bias as a difference in model performance for each gender. Instead of overall performance, they average over individual groups within a relationship (e.g., different individual occupations). This measure faces the challenge of generalizability as it relies on scraping a variety of existing text for different demographic groups.
 
\paragraph{Masked Language Model Predictions}
Several language representations are trained on the ability to predict masked words in text. 
CrowS-Pairs~\citep{nangia2020crows-pairs} and StereoSet~\citep{ nadeem2020stereoset} are datasets that use this property to expose and evaluate social biases learned with respect to 
different protected attributes. Both use crowdsourcing to obtain annotated sentence pairs, one of which is more stereotypical than the other for specific attributes (gender, socioeconomic status, etc). 
The evaluation metrics in both measures grade the model on its preference (through probabilities) for either the stereotypical or other sentence. 
Because these datasets permit crowdworkers to provide free-flowing text, the datasets are able to expand understandings of biases beyond a single stereotypical association across groups.
 
\paragraph{Text Classification (Occupations)}
\citet{De-artega2019bias} set up a measure for evaluating bias in text classification where the task is to predict a person's occupation given their biography.
The dataset contains short biographies crawled from online corpora using templates and removing sentences which contain occupation names.
Bias is evaluated by comparing results across different gender groups.
\citet{zhao2020gender} extend the original dataset to Spanish, French, and German. 
A challenge is equally scraping diverse data for different demographics, as reflected in the focus on binary gender for this measure.

\paragraph{Toxicity Detection} 
Toxic language ranges from more explicitly offensive forms (e.g., vulgar insults) to more subtle forms (e.g., microaggressions).
While toxicity detection aims to identify toxic language, existing works have found uneven detection of toxic language towards different groups. 
\citet{prabhakaran2019perturbation} show that there are varying levels of toxicity towards different names. 
\citet{dixon2018measuring} analyze biases in a toxicity classification model through the Wikipedia Talk Pages dataset as well as through a templated test set. 
Jigsaw~\citep{jigsaw} contains comments from the Civil Comments platform labeled with six types of toxicity (e.g., toxic, obscene, etc) and identity attributes (e.g., white, woman, etc). 
Along with this dataset, \citet{jigsaw} present a bias evaluation following that of~\citet{borkan2019nuanced} by comparing the AUC scores from different subgroups. 
Additionally, \citet{sap2020social} create a social bias inference corpus with toxicity labels 
and targeted group labels to understand the bias implications in languages.
These bias measures demonstrate 
that even tasks intended to detect harms may be biased.

\paragraph{Hate Speech Detection}
Hate speech detection is the task of identifying abusive language that is specifically directed towards a particular group.
To study biases in hate speech detection, many existing works have formulated different datasets and bias metrics.
\citet{davidson2017automated} and \citet{founta2018large} annotate Twitter datasets for hate speech detection.  \citet{blodgett2016demographic} provide a corpus of demographically-aligned text with geo-located messages based on Twitter. \citet{sap2019risk,xia2020demoting} use those datasets to show racial biases through a higher false positive rate for AAE, while \citet{davidson2019racial} use the dataset of \citet{blodgett2016demographic} for racial bias evaluation by comparing probabilities of tweets from different social groups being predicted as hate speech. \citet{davani2020fair} collect a dataset 
of comments from the Gab platform, but analyze biases by comparing a language model's log likelihood differences for constructed counterfactuals.  \citet{goldfarb2020intrinsic} add gender labels to the dataset from \citet{founta2018large} to analyze gender bias in hate speech detection, and further use \citet{basile2019semeval-2019}'s multilingual dataset to measure hate speech targeted at women and immigrants in English and Spanish. Similar to toxicity detection, most of these measures demonstrate the harm of online comments across demographic groups through a comparative score.

\paragraph{Bias Analyses without Complete Bias Measures}
There are other task-specific discussions of bias evaluations that do not propose specific bias measures.
For the task of common sense inference (incorporating common sense knowledge into model inference),
\citet{event2mind} analyze the intents of entities involved in an event, finding gender differences in the intents.
For named entity recognition, \citet{Mehrabi2020ManIT} discuss how models have different abilities to recognize male and female names as entities. 
For part-of-speech tagging, \citet{Munro2020DetectingIP} and \citet{Garimella2019WomensSR} find that state-of-the-art parsers
perform differently 
across genders,
failing to identify ``hers'' and ``theirs'' as pronouns but not ``his''. 
In addition, \citet{Mehrabi2021LawyersAD} and  \citet{rudinger2017social} demonstrate severe disparities in common sense knowledge and NLI datasets, respectively.


\subsection{Natural Language Generation}
We briefly describe some datasets and metrics used to evaluate biases in NLG tasks and refer readers to \citet{sheng2021societal} for a survey on common bias measures in Natural Language Generation.
For autocomplete generation, \citet{sheng2019woman} and \citet{huang2020reducing} both curate sets of prompts containing different demographic groups to prompt for inequalities in generated text. For the similar task of dialogue generation, \citet{liu2020does} construct a Twitter-based dataset with parallel context pairs between different groups, and
\citet{liu2020mitigating} rely on extracted conversation and movie datasets to evaluate gender biases.
Both works use various metrics such as sentiment, offensiveness, and the occurrence of specific words. For machine translation, the English WinoMT dataset \citep{stanovsky2019evaluating} is a widely used dataset for quantifying gender biases with bias metrics for translation typically rely on translation accuracy.

\section{Documenting Bias Measures}
\label{sec:appendix}
\subsection{Case Study \#1: Documentation for WinoBias \citep{zhao2018gender}}
\label{appendix:case-study-winobias}
\begin{enumerate}[leftmargin=*]

\item \textit{Motivation}

\begin{itemize}[leftmargin=*]
    \item \textbf{What is the stated definition of bias? How does this definition align with normative definitions of harm?}
    The paper defines gender bias in coreference resolution as the instance when a system associates pronouns to occupations that are dominated by the pronoun's associated gender more accurately than occupations \textit{not} dominated by that gender. While gendered associations with occupations are an instance of gender bias, such a definition does not capture gender bias in its entirety. The metric is defined to measure occupational perception of different genders, which is associated with representational harms.
    \item \textbf{What language and culture is the bias and measure most relevant to?} English language in the United States
     \item \textbf{If a demographic attribute is split into groups for measurement of bias, how many groups have been considered?} Gender binary (male and female) is considered in this measure.
    \item \textbf{What is the source of bias that is measured?} 
    The paper highlights two sources of gender bias: \textit{training data bias} and \textit{resource bias}. Training data used for coreference resolution systems are noted to have severe gender imbalance (over 80\% of entities headed by gendered pronouns are male). Pre-trained word embeddings which serve as an auxiliary resource for WinoBias \citep{zhao2018gender} have been shown to contain gender bias as well (“\textit{men}” is closer to “\textit{programmer}” than “\textit{woman}”). The paper also mentions a gender statistics corpus (i.e. \textit{Gender Lists}) as a resource that contains an uneven number of gendered contexts in which a noun phrase is observed.
    \item \textbf{What tasks or applications is this bias measure useful for?} 
    Since coreference resolution serves as an important step for many higher-level natural language understanding such as information extraction, document summarization, and question answering, this bias metric is useful for any of such tasks.
\end{itemize}

\item \textit{Composition and Collection Process}

\begin{itemize}[leftmargin=*]
    
    \item \textbf{Is the bias measure data scraped, generated, or produced some other way?} 
The data is created by the authors but the occupation list is collected from the U.S. Bureau of Labor Statistics. An advantage of this is that the profession categories come from an objective, rather than a biased, source as it is a government document. A disadvantage of this is that it is not comprehensive, and it is generated with the narrow view of only the United States.      

    \item \textbf{What are the limitations associated with method of data curation? How generalizable is this dataset?} 
    The data is limited because the occupations are collected from one source, and the source is specific to the United States. We expect that occupation titles and categories vary among different countries. Additionally, it is important to note that the statistics are constantly changing, and although the website that the data updates regularly, the dataset is static. This limits the relevance of the dataset as the world around it changes. 
    \item \textbf{Is the dataset at risk of causing harm through the particular selection of proxy attributes to represent demographic groups?} 
    Possibly---the dataset uses a limited set of occupations (curated from US-specific resources) and binary pronouns to represent different gender groups.
\end{itemize}

\item \textit{Bias Metrics}

\begin{itemize}[leftmargin=*]
    
    \item \textbf{How is the bias metric defined?} 
    It is defined as the absolute score difference between pro-stereotyped and anti-stereotyped conditions, where for pro-stereotypical condition, the gender pronoun is linked with the dominated profession and for anti-stereotypical vice versa.
    
    \item \textbf{Is it an absolute or relative evaluation?} 
    As it measures the bias through the difference between pro-stereotyped and anti-stereotyped conditions, it belongs to relative evaluation. Using a relative evaluation allows more flexibility for different models. 
    
    \item \textbf{Are there alternate or existing metrics this metric can or should be used with?} 
    WinoBias \citep{zhao2018gender} adapts the absolute difference of F1 to evaluate the gender bias. Since F1 score is a general metric to compare model performance, similar to the difference, the ratio could also be used to so disparity between to sets.
    
    
    \item \textbf{Are there other existing datasets or metrics to evaluate bias for the same task?}
    Yes, for coreference resolution task, there are also Gendered Ambiguous Pronouns (GAP)~\citep{webster2018mind} measuring the disparity incorrectly solving pronoun-name relationships for male and female genders, MAP~\citep{cao2020toward} (built on GAP beyond binary genders) and Winogender~\citep{rudinger2018gender} which also measures the relationship between gendered pronouns and occupations.
    
    
    \item \textbf{Can the metric imply an absolute absence of bias in a specific task or model?} 
    No, as discussed before, this metric only focuses on entities with 40 occupations in limited sentence templates. Even if the absolute difference doesn’t show much inequalities, there could still be biases in the model.
    
\end{itemize}

\end{enumerate}
\subsection{Case Study \#2: Documentation for Regard \citep{sheng2019woman}}
\label{appendix:case-study-regard}
\begin{enumerate}[leftmargin=*]
    
\item \textit{Motivation}

\begin{itemize}[leftmargin=*]
    \item \textbf{What is the stated definition of bias? How does this definition align with normative definitions of harm?}
    The authors do not provide an explicit definition of bias, but define bias in terms of the metric of \textit{regard} (i.e., social perception) towards a demographic, which can be negative, neutral or positive.
    Since this metric is defined to measure social perception, it is aligned with definitions of representational harms, e.g., negative stereotypes, denigrations.
    \item \textbf{What is the source of bias that is measured?} 
    It is difficult to pinpoint the exact sources of biases from the probing experiments run by \citet{sheng2019woman} on GPT-2 and the 1 Billion Word Language Model, though we can form hypotheses.
    While the One Billion Word Benchmark dataset is publicly available for analysis, the exact dataset used to train GPT-2 can probably only be approximated at best.
    However, we know that GPT-2 was trained on Web data, including from Web sources such as Reddit, which the authors mention as a likely source of biases.
    The 1 Billion Word Language Model was trained on news data, and \citet{sheng2019woman} find less biased results from this model.
    There could also be non-data related biases (e.g., depending on features in the model architecture and training procedure), though more studies need to be done here.
    \item \textbf{What tasks or applications is this bias measure useful for?} 
    The metric of regard is useful for applications for continuation generation tasks \citep{sheng2021societal}, e.g., when a system takes an input prompt and generates text in a mostly unconstrained manner.
    In other words, this metric could also be useful for dialogue generation, chat bots, virtual assistants, and creative generation applications, in addition to language models.
\end{itemize}

\item \textit{Composition and Collection Process}

\begin{itemize}[leftmargin=*]
    \item \textbf{Is the bias measure data scraped, generated, or produced some other way?} 
    The data used as input prompts to probe for biases are generated from templates.
    For example, \textit{``XYZ worked as''}, \textit{``XYZ earned money by''}, etc.
    These templates allow for a controlled probing of inequalities in specific contexts related to occupations and respect.
    The disadvantages are that templates can be time-consuming to manually construct (\citet{sheng2019woman} only use 10 templates) and may not be representative or comprehensive of all the ways that similar content could be phrased.
    Additionally, the templates could be biased towards the syntactic and semantic inclinations of the template creators, which may or may not align with those the model is used to seeing.
    \item \textbf{What are the limitations associated with method of data curation? How generalizable is this dataset?} 
    These templates are generalizeable to other demographic surface forms not mentioned in original work.
    Although conceptually these templates can be extended to probe biases in other contexts (e.g., contexts likely to lead to negative religious or ethnic stereotypes), manually creating these contexts is slow and likely not comprehensive.
    While these templates could also be translated to other languages, relying on automatic translations could result in unnatural phrasings, while manual translations are more time-consuming.
\end{itemize}

\item \textit{Bias Metrics}
\begin{itemize}[leftmargin=*]
    \item \textbf{How is the bias metric defined?}
    \citet{sheng2019woman} define the metric of regard (social perception) towards a demographic group. Possible values are negative, neutral, or positive.
    \item \textbf{Is it an absolute or relative evaluation?}
    The authors have formatted the comparison of regard scores across demographics as a relative evaluation. 
    Using a relative evaluation allows more flexibility for different analyses. 
    \item \textbf{Are there alternate or existing metrics this metric can or should be used with?} 
    \citet{sheng2019woman} show in their study (Table 5) that the metrics of sentiment and regard can be well-correlated for some types of prompts yet greatly differ for other types.
    They conclude that it could be useful to report both sentiment and regard.
    \item \textbf{Are there other existing datasets or metrics to evaluate bias for the same task?}
    At the time of publication, there were perhaps limited proposed alternatives for evaluating biases from language models, though there are now other options.
    \citet{huang2020reducing} present 730 manually curated templates to probe for sentiment differences across countries, occupations, and genders in language models.
    There are also other bias measures for language models that rely on sentiment \citep{groenwold2020investigating,shwartz2020you}.
    Both \citet{sheng2019woman} and \citet{huang2020reducing} construct manual prompts to test for biases towards demographics mentioned in the input.
    Additionally, \citet{groenwold2020investigating} evaluate for similar biases in language models towards \emph{people who produce the text} \citep{sheng2021societal}.
    Combining all these bias measures would provide a more comprehensive analysis.
\item \textbf{Can the metric imply an absolute absence of bias in a specific task or model?} 
    No, as discussed in earlier answers, the limited templates (both in number and in syntactic/semantic diversity) mean that even if the regard scores do not show inequalities, there could still be biases in the model.
    Also, since the authors use a regard classifier to feasibly automatically label a large number of samples, there could also be biases from the classifier itself.
    Even human evaluations of regard could be influenced by human biases.
\end{itemize}
\end{enumerate}


\end{document}